\documentclass[letterpaper]{article} 
\usepackage{aaai25}  
\usepackage{times}  
\usepackage{helvet}  
\usepackage{courier}  
\usepackage[hyphens]{url}  
\usepackage{graphicx} 
\usepackage{natbib}  
\usepackage{caption} 
\usepackage{algorithm}
\usepackage{algorithmic}
\usepackage{color}
\usepackage{amsmath}
\usepackage{amssymb}
\usepackage{multirow}
\usepackage{newfloat}
\usepackage{listings}
\usepackage{newfloat}
\usepackage{listings}
\DeclareCaptionStyle{ruled}{labelfont=normalfont,labelsep=colon,strut=off} 
\floatstyle{ruled}
\newfloat{listing}{tb}{lst}{}
\floatname{listing}{Listing}
\title{Pilot: Building the Federated Multimodal Instruction Tuning Framework}
\author{
    Baochen Xiong\textsuperscript{1,2,3},
    Xiaoshan Yang\textsuperscript{1,2,3},
    Yaguang Song\textsuperscript{2},
    Yaowei Wang\textsuperscript{2,4}, 
    Changsheng Xu\textsuperscript{1,2,3\thanks{Corresponding author.}}
}
\affiliations{
    \textsuperscript{\rm 1}MAIS, Institute of Automation, Chinese Academy of Sciences\\
    \textsuperscript{\rm 2}Pengcheng Laboratory\\
    \textsuperscript{\rm 3}School of Artificial Intelligence, University of Chinese Academy of Sciences (UCAS)\\
    \textsuperscript{\rm 4}Harbin Institute of Technology (Shenzhen)\\

    {xiongbaochen2022@ia.ac.cn, \{xiaoshan.yang, csxu\}@nlpr.ia.ac.cn, \{songyg01, wangyw\}@pcl.ac.cn}
}

\usepackage{bibentry}
\begin{document}

\maketitle

\begin{abstract}
In this paper, we explore a novel federated multimodal instruction tuning task(FedMIT),
which is significant for collaboratively fine-tuning MLLMs on different types of multimodal instruction data on distributed devices.
To solve the new task, we propose a federated multimodal instruction tuning framework(Pilot).
Our framework integrates two stages of ``adapter on adapter” into the connector of the vision encoder and the LLM.
In stage 1, we extract task-specific features and client-specific features from visual information.
In stage 2, we build the cross-task Mixture-of-Adapters(CT-MoA) module to perform cross-task interaction.
Each client can not only capture personalized information of local data and learn task-related multimodal information, but also learn general knowledge from other tasks.
In addition, we introduce an adaptive parameter aggregation strategy for text training parameters, which optimizes parameter aggregation by calculating weights based on the euclidean distance between parameters, so that parameter aggregation can benefit from positive effects to the greatest extent while effectively reducing negative effects.
Our framework can collaboratively exploit distributed data from different local clients to learn cross-task knowledge without being affected by the task heterogeneity during instruction tuning.
The effectiveness of our method is verified in two different cross-task scenarios.

\end{abstract}

%

\section{Introduction}
The emergence of Multimodal Large Language Models (MLLMs)~\cite{li2023blip,liu2024visual,driess2023palm,instructblip} has significantly advanced the field of artificial intelligence.
MLLMs have shown excellent ability in processing and integrating various modalities information (especially text and image), and has achieved remarkable performance in tasks such as text generation, machine translation and question answering.
Enhancing the zero-shot generalization ability of MLLMs on novel multimodal tasks is a key goal driving its development.
Multimodal instruction tuning has been shown to be highly effective in improving the zero-shot generalization ability of models to unseen multimodal problems~\cite{xu2022multiinstruct,ye2023mplug,sun2024umie,chen2024visual,xiao2024oneref}.

\begin{figure}[t]
\centering
    \includegraphics[width=1\linewidth]{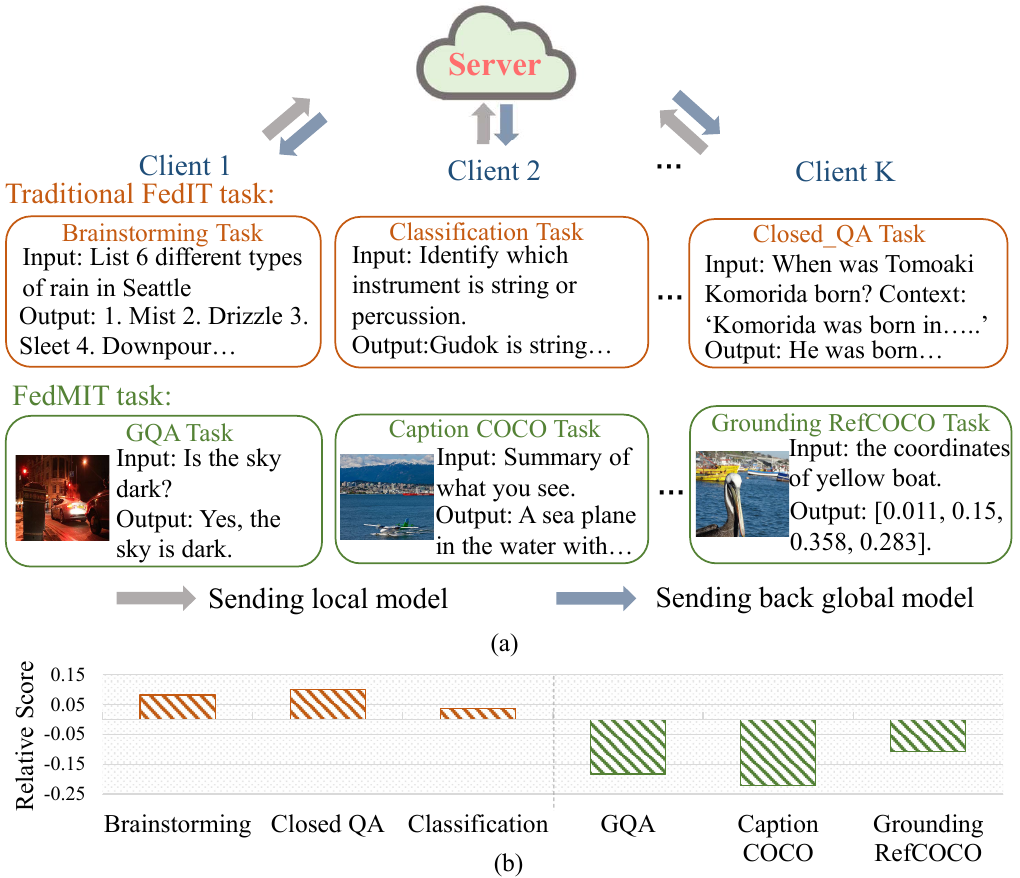} 
    \caption{(a) Comparison of the FedIT task with the FedMIT task. 
    (b) Compare the relative scores of traditional FedIT methods on the two tasks.
    } 
    \vspace{-1.3em}
    \label{motivation}
\end{figure}

Current instruction tuning methods are usually implemented by centralized learning paradigm, where one central party collects a substantial amount of data to train the model, which may lead to privacy security issues.
%
%
Federated Learning (FL) is currently the primary framework for distributed training of models while maintaining data security and privacy.
\citeauthor{mcmahan2017communication}~\shortcite{mcmahan2017communication} proposed the first federated learning algorithm, FedAvg, which aggregates model gradients from local clients to a central server without data sharing.
Afterwards, FL is also successfully used for instruction tuning of pre-trained models~\cite{zhang2024towards,ye2024openfedllm},
called Federated Instruction Tuning (FedIT).
These frameworks can effectively leverage locally private instruction-following data.
Although existing FedIT approaches have made important progress, they only consider the case where different clients perform different instruction-based natural language processing(NLP) tasks, and there is little exploration in the multimodal field.
The integration of vision and language plays a crucial role in machine perception and understanding of the world.
Language is the basis of cognitive processing, while vision provides the necessary sensory information~\cite{xu2024libra, xiong2024modality}.
In this context, we attempt to transfer the success of FedIT in multimodal tasks.
Therefore, in this paper, we explore a novel Federated Multimodal Instruction Tuning task (FedMIT).
%

%
%

%
Compared with traditional FedIT, FedMIT focuses on the client containing different multimodal instruction tuning tasks (e.g., visual question answering and image captioning), as shown in Figure~\ref{motivation}(a).
In our preliminary study,
we first distributed different task instruction data to each client, including ``Caption (e.g., COCO~\cite{lin2014microsoft})" for image description, ``VQA (e.g., GQA~\cite{hudson2019gqa})" for visual question answering, and ``Grounding (e.g., RefCOCO~\cite{kazemzadeh2014referitgame})" for visual localization.
%
Then, we directly apply the representative method Shepherd~\cite{zhang2024towards} in the FedIT task to the FedMIT task.
%
%
Figure~\ref{motivation}(b) shows the relative scores of Shepherd on two tasks, where the relative scores are the comparison scores between the results of the Shepherd method and the local training results.
%
%
The results show that the traditional FedIT method does not perform well on the FedMIT task.
%
Since the diversity of multimodal tasks greatly increases the heterogeneity between clients,
we believe that traditional FedIT methods cannot adequately address this kind of task heterogeneity.
Compared with the traditional FedIT task, each client in FedMIT task not only needs to capture the personalized information of local data and task-related multimodal information, but also needs to be able to accommodate the differences between different tasks to avoid parameter conflicts.
This requires the model to maintain its understanding of the task and local data while also being able to learn general knowledge from other tasks to improve model performance and cross-task ability.
Inspired by these observations, we introduce the Federated Multimodal Instruction Tuning framework: Pilot.
%
%
Our framework integrates two stages of ``adapter on adapter" into the connector of the vision encoder and the LLM, and adopts an adaptive parameter aggregation strategy for the training parameters of the LLM.
First we introduce the two-stage training of the client.
\textbf{Stage 1}: Task-specific feature mining.
We hope to extract client-specific and task-specific features from the client's visual information.
%
%
We propose task-specific adapter to extract task-specific visual features that is only important for one task, and client-specific adapter to extract specific visual features of the client’s unique data distribution.
%
%
%
To encourage the client-specific adapter to produce features that are more refined than the task-specific adapter, a difference loss is used to ensure the orthogonality of their output features.
%
%
%
\textbf{Stage II}: Cross-task visual interaction.
We integrate the Cross-task Mixture-of-Adapters(CT-MoA) module with the task-specific adapter.
By interacting with the server, each adapter in CT-MoA is initialized from the task-specific adapter of the corresponding task.
%
We hope that the CT-MOA module can learn general knowledge from other tasks to improve model performance and cross-task capabilities.
Therefore, we added cross-task adapters to the task-specific adapters of other tasks in CT-MOA, where cross-task adapter on other task-specific adapter.
Cross-task adapter aims to extract cross-task collaboration visual features.
%
%
In addition, the CT-MOA module also contains a router that selects adapters during the stage II with auxiliary losses on the router to maintain a balanced loading of adapters.
Considering the computation and communication requirements, we adopt text-adapter-based parameter-efficient tuning techniques to train LLM to reduce the amount of trainable parameters on each device.
%
%
For the server side, it collects all client visual and text training parameters.
For visual training parameters, we adopt the task-aware aggregation strategy.
For text training parameters, we introduce adaptive parameter aggregation, which optimizes parameter aggregation by calculating weights based on the euclidean distance between parameters,
so that parameter aggregation can benefit from positive effects to the greatest extent while effectively reducing negative effects.
%
%
Combining federated optimization with two-stage local updates, our framework can collaboratively exploit distributed data from different local clients to learn cross-task knowledge without being affected by the task heterogeneity during instruction tuning.
%
%

Our contributions are summarized as follows: 
(1) We propose to explore a new task of federated multimodal instruction tuning, which is  significant for collaboratively fine-tuning MLLMs on different types of multimodal instruction data on distributed devices.
(2) To solve the new task, we propose a Federated Multimodal Instruction Tuning framework(Pilot).
%
%
Our framework builds two stages ``adapter on adapter" strategy.
In stage 1, the model extracts client-specific and task-specific features,
and in stage 2, 
we construct CT-MOA modules to learn cross-task interactions.
We adopt an adaptive aggregation strategy for the LLM training parameters.
With the above approach, our method can learn cross-task knowledge without being affected by task heterogeneity during instruction tuning.
(3) We verify the effectiveness of Pilot on the state-of-the-art LLaVA~\cite{liu2024visual} in two different cross-task scenarios.

\begin{figure*}[t]\label{framework}
\centering
    \includegraphics[width=0.95\linewidth]{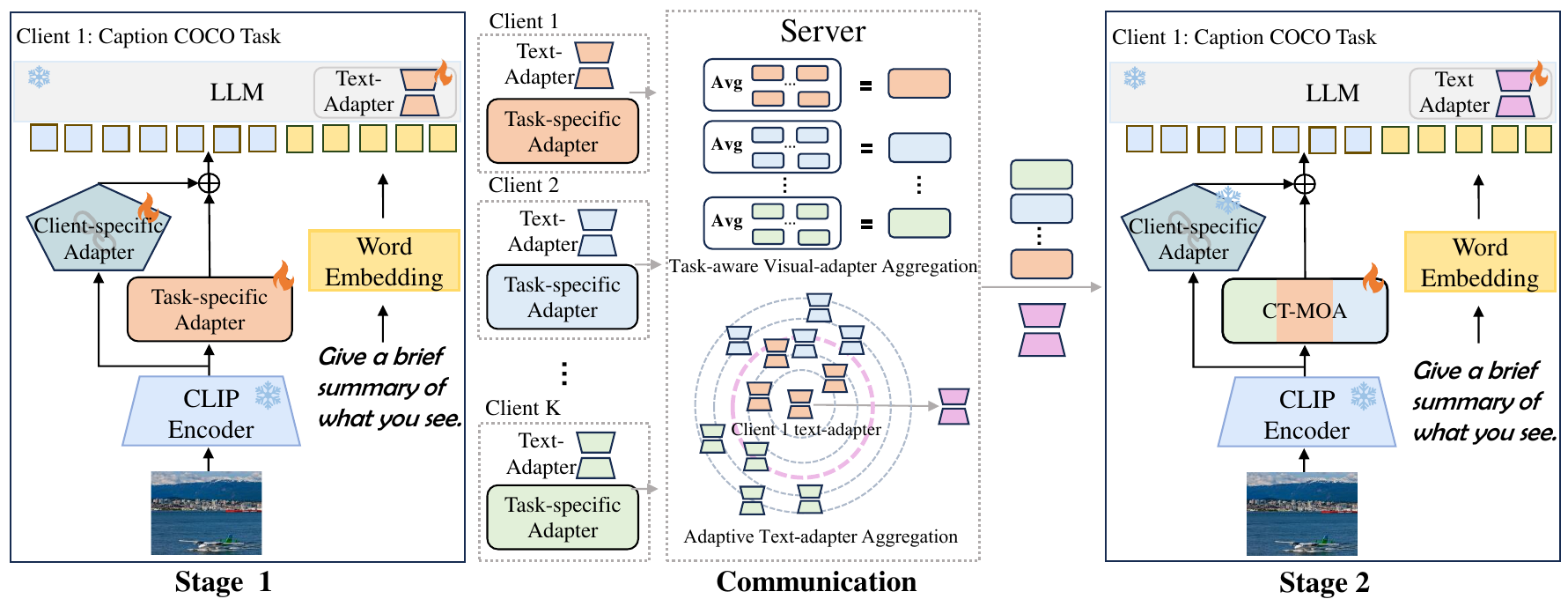}
    \caption{
    Overview of the proposed Federated Multimodal Instruction Tuning Framework (Pilot). 
}\label{fig:framework}
\vspace{-1em}
\end{figure*}

\section{Related Work}
\subsection{Federated Learning}
The earliest FL algorithm is FedAvg~\cite{mcmahan2017communication}, which builds the global model by averaging the local updates obtained by Stochastic Gradient Descent (SGD)~\cite{gorbunov2021local}.
However, FedAvg inevitably experiences performance degradation on non-IID data~\cite{yang2024cross,xiong2023client}.
To deal with this problem,
FedProx~\cite{li2020federated} adds a proximal term to local targets to minimize the distance between the global model and the local model for non-IID data.
PerAvg~\cite{fallah2020personalized} uses popular meta-learning framework MAML~\cite{finn2017model},
which allows each client to quickly adapt to local data by finding a suitable initialization.
%
%
TAPPFL~\cite{arevalo2024task} designs a task-agnostic and provably privacy-preserving federated learning framework.
FedTGP~\cite{zhang2024fedtgp} uses adaptive-margin-enhanced contrastive learning to learn trainable global prototypes on the server to solve the model heterogeneity problem.
FedLPS~\cite{jia2024fedlps} uses an adaptive channel model pruning method to enable clients to participate in federated learning training with heterogeneous tasks.
However, the above methods only consider the heterogeneity on single-modal clients.
In contrast, our method focuses on addressing the heterogeneity of cross-task instruction data in the MLLM-based federated multimodal instruction tuning task.

\subsection{Multimodal Instruction Tuning}
Instruction tuning~\cite{wei2021finetuned,xiao2024hivg} significantly improves the generalization ability of large language models to unseen tasks based on natural language instructions.
With the advent of MLLMs, the scope of instruction tuning has been expanded to include multimodal and visual tasks.
MiniGPT-4~\cite{zhu2023minigpt} and LLaVA~\cite{liu2024visual} keep the visual encoder frozen and adjust the language model, extending instruction tuning to multimodal.
InstructBLIP~\cite{instructblip} enhances zero-shot capabilities by performing instruction tuning on multiple datasets. 
Shikra~\cite{chen2023shikra} extend MLLMs to visual grounded tasks using instructions with bounding box coordinates.
mPLUG-Owl~\cite{ye2023mplug} fine-tunes the visual and text encoders in two stages.
EProj~\cite{he2023continual} address the catastrophic forgetting of the model in continuous instruction tuning. 
MixLoRA~\cite{shen2024multimodal} uses conditional mixed LoRA to address the task interference caused by LoRA during multimodal instruction tuning.
Although these studies demonstrate excellent zero-shot capabilities, they still suffer from the risk of privacy leakage.

\subsection{Federated Instruction Tuning}
There has been some research in federated learning based on Parameter-Efficient Fine-tuning(PEFT)~\cite{shi2023clip,su2024federated,chen2024disentanglement}.
FedCLIP~\cite{lu2023fedclip} and pFedprompt~\cite{guo2023pfedprompt} finetune the CLIP model~\cite{radford2021learning} by adapting and conveying only small number of learnable prompts.
FedDISC~\cite{yang2024exploring} first integrated the pre-trained diffusion model~\cite{dhariwal2021diffusion} into the FL framework.
FedDAT~\cite{chen2024feddat}utilizes a dual-adapter teacher architecture to address the efficient fine-tuning of parameters of multimodal base models in heterogeneous federated learning.
However, the application of instruction tuning in FL has not been fully explored.
Federated instruction tuning~\cite{zhang2024towards} provides a simple and effective method for supporting distributed client privacy-preserving instruction tuning through the FL protocol.
Fatellm~\cite{fan2023fate} and OpenFedLLM~\cite{ye2024openfedllm} built a concise and easy-to-research framework for fine-tuning federal instructions.
Recently, FedDPA~\cite{yang2024dual} explored the problem of heterogeneity in NLP tasks.
In our work, we attempt to address the unexplored task of multimodal instruction tuning for MLLMs in federated learning.

\section{Methodology}
\subsection{Problem Definition}
%
We assume that there is a set of multimodal instruction tuning data $\mathcal{D}=\{(\mathcal{D}_k,{t}_k)\}_{k=1}^K$ from $K$ clients, where $\mathcal{D}_k=\{{X}_{i,k}^{v},{\bf x}_{i,k}^{ins}, {\bf y}_{i,k}^{ans}\}_{i=1}^{n_k}$ is the set of $n_k$ data pairs on the $k$-th client.
The total number of data pairs is $n=\sum_{k=1}^Kn_k$.
${X}^{v}, {\bf x}^{ins}$ and ${\bf y}^{ans}$ indicate the image, instruction tokens and answer tokens, respectively.
${t}_k\in \{1,...,T\}$ denotes the task type of the $k$-th client.
$T$ is the total number of tasks and $T\leqslant K$.

In the FedMIT task, all clients obtain the pre-trained MLLM from the server. 
Generally, MLLMs contain a visual encoder $f$, a connector $\psi$, and LLM $L$. 
Specifically, for a given input image ${X}^{v}$, the visual encoder extracts visual features $H^v$ = $f$(${X}^{v}$).
%
%
%
%
The connector is used to align the visual encoder with the LLM.
Connector transforms $H^v$ into a language embedding tokens ${\bf x}^{img}\in \mathbb{R}^{N \times C}$, effectively facilitating the integration of multimodal information within the LLM framework, where $N$ is the number of tokens and $C$ is the hidden size.
\begin{equation}
{\bf x}^{img} = \psi(H^v) \text {, with } H^v=f({X}^{v}).
\end{equation}
Finally, we input ${\bf x}^{img}$ and ${\bf x}^{ins}$ into the LLM to generate response.
The FedMIT task aims to perform instruction tuning in a distributed manner, where each client can increment the local model by leveraging cross-task knowledge learned on other clients without the need for centralized data collection. 
The overall optimization goal is defined as follows:
\begin{equation}
\sum_{k=1}^{K}{\frac{n_k}{n}}\sum_{i=1}^{n_k}{\frac{1}{n_k}}\sum_{j=1}^L -\log p_\theta\left({\bf y}^{j,ans}_{i,k} \mid {\bf x}^{img}_{i,k}, {\bf x}^{ins}_{i,k}, {\bf y}^{<j,ans}_{i,k}\right)
\end{equation}
where $L$ represents the answer length, $\theta$ is the trainable parameters.
%
${\bf y}^{j,ans}_{i,k}$ indicates the $j$-th answer token and ${\bf y}^{<j,ans}_{i,k}$ indicates all answer tokens before the index $j$.

\subsection{Federated Multimodal Instruction Tuning Framework}
FedMIT task has greater heterogeneity between tasks, and traditional FedIT methods cannot be used directly to solve the problem.
%
%
As shown in Figure~\ref{fig:framework}, we propose the Federated Multimodal Instruction Tuning framework (Pilot) to address the task heterogeneity between clients.
In our framework, we integrate a two-stage ``adapter on adapter" method into the connector of the visual encoder and LLM.
In stage 1, we extract task-specific features and client-specific features from visual information.
Through federated aggregation, in stage 2, we build a CT-MOA module to perform cross-task interaction.
We hope that each client not only needs to capture personalized information from local data and learn task-related multimodal information, but also needs to be able to learn general knowledge from other tasks to improve model performance and cross-task capabilities.

\textbf{Stage 1: Task-specific Feature Mining.}
At stage 1, we hope to extract client-specific and task-specific features from the client’s visual information.
%
We propose task-specific adapter ${\psi}^t$ to extract task-specific visual features that is only important for one task, and client-specific adapter ${\psi}^s$ to extract specific visual features of the client’s unique data distribution.
We define these two adapters as two-layer of perceptrons.
%
%
Finally, the image tokens represent:
${\bf x}^{img} = {\bf x}^{t} + {\bf x}^{s}$, where ${\bf x}^{t} = {\psi}^{t}(H^v)\in \mathbb{R}^{N \times C}$, ${\bf x}^{s} = {\psi}^{s}(H^v)\in \mathbb{R}^{N \times C}$.
To encourage the client-specific adapter to produce features that are more refined than the task-specific adapter, a difference loss is used to ensure the orthogonality of their output features.
Inspired by the domain separation network~\cite{bousmalis2016domain}, we adopt a soft subspace orthogonality constraint:
\begin{align}
\mathcal{L}_{d}=\|{{\bf x}^{t}}^{\top}{\bf x}^{s}\|_F^2,
\end{align}
where $\|\cdot\|_F$ is Frobenius norm.
Hence, the Stage 1 total loss is
\begin{align}
\mathcal{L}=\mathcal{L}_{ce} + \lambda_0\mathcal{L}_{d},
\end{align}
where $\mathcal{L}_{ce}$ represents the language modeling loss, which computes the cross-entropy of next-token predictions.
$\lambda_0$ denote coefficients for difference loss.
After stage 1, each client sends the LLM training parameters text-adapter $\Theta_{l}$ and task-specific adapter $\Theta_{a}$ parameters to the server.


\begin{figure}[t]
\centering
    \includegraphics[width=0.9\linewidth]{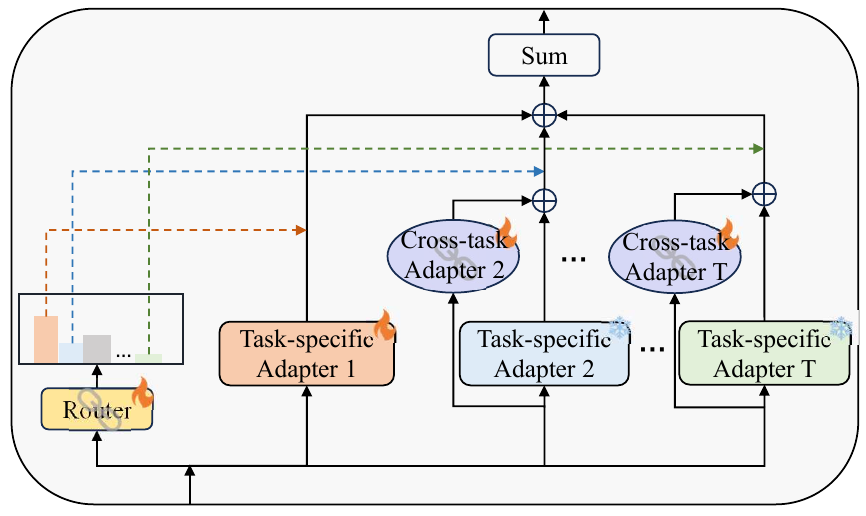}
    \caption{Cross-task Mixture-of-Adapters (CT-MoA).} 
    \label{MOA}
    \vspace{-1.0em}
\end{figure}

\textbf{Stage 2: Cross-task Visual Interaction.}
At stage 2, the client obtains the updated task-specific adapters for $T$ tasks from the server.
In other words, each client obtains $(T-$$1)$ task-specific adapters of other tasks in addition to its local task-specific adapter.
To learn general knowledge from other tasks to improve model performance and cross-task capabilities,
we integrate the local task-specific adapter with other task-specific adapters to construct a Cross-task Mixture-of-Adapters (CT-MoA) module, as shown in Figure~\ref{MOA}.
In the CT-MoA module, we assume that the local task-specific adapter index is 1 (${\psi}_1^t$), and there are $T$ adapters in total.
However, directly loading task-specific adapters of distinct tasks may cause the model of the current task to be unable to fully utilize these adapters due to task heterogeneity.
Therefore, we added cross-task adapters to the task-specific adapters of other tasks in CT-MOA.
For task-specific adapters ${\psi}_i^t, i=2,...,T$, we add cross-task adapter ${\psi}^c$ on it
to alleviate the discrepancy between other task-specific adapters and the local task-specific adapter due to task heterogeneity.
Cross-task adapter aims to extract cross-task collaboration knowledge.
It has the same structure as the task-specific adapter and is initialized by the local task-specific adapter ${\psi}_1^t$ parameters.
In addition, the CT-MOA module includes a router for predicting the probability of selecting and activating each adapter from $T$ total task-specific adapters.
The router network($\phi$) has a linear layer for calculating the normalized weight matrix using $H^v$ for voting, and producting $\mathcal{P}$:
\begin{equation}
\mathcal{P}=\operatorname{Softmax}(\phi(H^v)) \in \mathbb{R}^{T}.
\end{equation}
Finally, the output of CT-MOA module is expressed as:
\begin{equation}
{\bf x}^{t} =\mathcal{P}[1]{\psi}_1^t(H^v)+\sum_{i=2}^T\mathcal{P}[i]\left({\psi}_i^t(H^v)+ {\psi}_i^c(H^v)\right).
\end{equation}
Following~\cite{zoph2022st}, we adopt an auxiliary losses based on the language modeling cross-entropy loss to maintain a load balance between task-specific adapters in the Cross-task MoA module.
The auxiliary losses comprise load balancing loss and router z-loss.
The load balancing loss is defined as:
\begin{equation}
\begin{aligned}
\mathcal{L}_{b}=T \sum_{i=1}^T \mathcal{D}_i \mathcal{R}_i,
\end{aligned}
\end{equation}
where $\mathcal{D}_i=\frac{1}{N} \sum_{j=1}^N 1\left\{\operatorname{argmax} (\phi(H^v[j]))=i\right\}$
represents the proportion of tokens distributed to adapter $i$, $1\{\cdot\}$ is an indicator function.
While $\mathcal{R}_i=\frac{1}{N} \sum_{j=1}^N \phi_{i}(H^v[j])$ denotes the proportion of the probability assigned to adapter $i$.
$\phi_{i}(H^v[j])$ is the probability of routing token $j$ to adapter $i$.
The router z-loss is defined as:
\begin{equation}
\mathcal{L}_{z} =\frac{1}{N} \sum_{i=1}^N\left(\log \sum_{j=1}^T e^{g_j^{(i)}}\right)^2,
\end{equation}
where $g \in \mathbb{R}^{N \times T}$ are the logits obtained by the router.
Hence, the Stage 2 total loss is:
\begin{align}
\mathcal{L}=\mathcal{L}_{ce} + \lambda_1\mathcal{L}_{b} + \lambda_2\mathcal{L}_{z},
\end{align}
$\lambda_1$ and $\lambda_2$ denote coefficients for load balancing loss and router z-loss.
During stage 2 training, the model only updates the local task-specific adapter, the cross-task adapter, and the text-adapter.

\subsection{Federated Optimization}
The proposed method is collaboratively optimized via local two stages instruction tuning and global aggregation.
We train the Pilot with two alternate steps of local update and global aggregation for $R$ rounds.
In each round, the client receives the global model and updates the local model $E$ epochs on the local data.
In the first round, the server sends the base MLLM model to each client.
Each client performs local instruction tuning of stage 1, then send the task-specific adapter and text-adapter parameters to the server.
After global aggregation, the server sends $T$ task-specific adapters and text-adapter parameters back to the client.
%
Each client updates the model architecture and performs local instruction tuning of stage 2, and finally send the local task-specific adapter and text-adapter parameters to the server.
%
%

\textbf{Task-aware Visual-adapter Aggregation.}
After the server collects the task-specific adapter parameters of all clients, we adopt a task-aware aggregation method.
Task-specific adapter mainly learns task-specific visual features and has the general understanding ability of this task.
Therefore, the parameters learned in one client can also be shared by clients of the same task.
We use weighted average to aggregate the task-specific adapter parameters $\Theta_a$ of the same task:

\begin{align}
\bar{\Theta}^{t}_a=\sum_{k\in \mathcal{K}_t}\frac{n_k}{\sum\limits_{{k'}\in \mathcal{K}_t}n_{k'}}{\Theta}^{t}_{a,k},  t=1,...,T,
\end{align}
where $\mathcal{K}_t$ is a set of clients with task type $t$.
Finally, we obtain $T$ number of task-specific adapters and send them to each client.

\textbf{Adaptive Text-adapter Aggregation.}
For all collected text-adapter parameters, compared with the fully weighted aggregation method used in the traditional FedIT method, we expect that the text-adapter parameters of each client can benefit from the positive impact to the greatest extent during the aggregation process, while effectively reducing the negative impact.
Therefore, we propose an adaptive text-adapter aggregation method.
We first calculate the Euclidean distance between each text-adapter parameter and the other $(K-$$1)$ text-adapter parameters.
Then select the nearest $M$ points for weighted aggregation based on the distance.
For example, text-adapter parameters $\Theta_{l,k}$ for client $k$:
\begin{equation}
d_{k,i} = E(\Theta_{l,k}, \Theta_{l,i}), \text { for } i \in\{1,2, \ldots, K\} \backslash\{k\},
\end{equation}
where $E$(·) is the Euclidean distance formula.
We obtain the distance $D_k \in \mathbb{R}^{K-1}$ between the $\Theta_{l,k}$ and other text-adapter parameters, 
and then we select the Top-$M$ closest parameters for weighted aggregation:
\begin{align}
\bar{\Theta}_{l,k}=\frac{n_k}{n_k+\sum\limits_{{k'}\in \mathcal{K}_m}n_{k'}}\Theta_{l,k}+
\sum_{k'\in \mathcal{K}_m}\frac{n_{k'}\cdot{w_{k'}}}{n_k+\sum\limits_{{k'}\in \mathcal{K}_m}n_{k'}}{\Theta}_{l,k'},
\end{align}
where $w_{k'}=\frac{1 / d_{k,k'}}{\sum_{k'\in \mathcal{K}_m} 1/d_{k,k'}}$,
$\mathcal{K}_m$ is the set of $M$ clients closest to parameter $\Theta_{l,k}$.

\section{Experiment}
\subsection{Experimental Setups}
Now, we introduce how to construct the federated multimodal instruction tuning task scenario.
To ensure the diversity of instruction tuning datasets,
we collect various publicly available and commonly used visual-language datasets.
These instruction tuning datasets cover a wide range of tasks, including knowledge-based image question answering, image question answering reading comprehension, and optical character recognition VQA.
The selected datasets include ScienceQA~\cite{lu2022learn}, GQA~\cite{hudson2019gqa}, and OCRVQA~\cite{mishra2019ocr}.
However, we observe that these datasets are limited to traditional QA tasks in visual-language tasks.
Therefore, to enrich the diversity of tasks, we introduce the image caption dataset COCO~\cite{lin2014microsoft} for image description task and the grounding dataset RefCOCO~\cite{kazemzadeh2014referitgame} for visual localization task. 
For all the above datasets, we construct two different federated instruction tuning scenarios.
\textbf{FL-oriented visual understanding scenario}: We use the GQA, COCO, and RefCOCO datasets.
We randomly divide each dataset into 3 subsets, and each subset is regarded as a client.
Then, we obtain a FedMIT task scenario with 9 clients and 3 different visual understanding tasks(9-client, 3-task).
\textbf{FL-oriented general VQA scenario}: We use the ScienceQA, GQA, and OCRVQA datasets.
Similar to the above operations, we obtain a FedMIT task scenario with 9 clients and 3 different VQA tasks(9-client, 3-task).
For more data information, please refer to the supplementary materials.

\begin{table*}[t]
\setlength{\tabcolsep}{8.5pt}
\centering
\renewcommand{\arraystretch}{1.1}
\small
\captionsetup{font={normalsize}}
\begin{tabular}{c|ccc|ccc|ccc}
\hline
\multirow{2}{*}{\textbf{Methods}} & \multicolumn{3}{c|}{\textbf{GQA}} & \multicolumn{3}{c|}{\textbf{Caption COCO}} & \multicolumn{3}{c}{\textbf{Grounding RefCOCO}} \\ \cline{2-10} 
 & Client 1 & Client 2 & Client 3 & Client 4 & Client 5 & Client 6 & Client 7 & Client 8 & Client 9 \\ \hline
Centralized training & \multicolumn{3}{c|}{Centralized test result: 52.8} & \multicolumn{3}{c|}{Centralized test result: 130.3} & \multicolumn{3}{c}{Centralized test result: 65.3} \\ \hline
Local training & 48.8 & 47.5 & 48.4 & 117.6 & 122.3 & 120.9 & 29.1 & 35.1 & 31.9 \\ \hline
FedAvg & 47.2 & 46.6 & 46.4 & 86.5 & 102.5 & 104.2 & 25.4 & 30.2 & 32.4 \\
FedProx & 48.1 & 46.4 & 47.8 & 98.5 & 101.2 & 113.2 & 35.8 & 45.5 & 39.6 \\
FedAdam & 50.8 & 48.8 & \textbf{50.5} & 117.3 & 119.2 & 122.0 & 47.5 & 45.9 & 45.1 \\
FedDPA & 49.3 & 48.2 & 49.3 & 112.2 & 121.4 & 113.6 & 48.2 & 47.3 & 46.2 \\
Shepherd & 43.8 & 39.6 & 44.7 & 75.1 & 86.2 & 101.2 & 24.1 & 30.7 & 28.5 \\ \hline
\textbf{Pilot} & \textbf{51.4} & \textbf{49.7} & 50.2 & \textbf{120.3} & \textbf{126.4} & \textbf{125.1} & \textbf{49.6} & \textbf{52.4} & \textbf{50.9} \\ \hline
\end{tabular}
\caption{Comparison with state-of-the-art methods under FL-oriented visual understanding scenario.}
\vspace{-0.5em}
\label{result1}
\end{table*}

\begin{table*}[t]
\setlength{\tabcolsep}{8.5pt}
\centering
\renewcommand{\arraystretch}{1.1}
\small
\captionsetup{font={normalsize}}
\begin{tabular}{c|ccc|ccc|ccc}
\hline
\multirow{2}{*}{\textbf{Methods}} & \multicolumn{3}{c|}{\textbf{GQA}} & \multicolumn{3}{c|}{\textbf{ScienceQA}} & \multicolumn{3}{c}{\textbf{OCRVQA}} \\ \cline{2-10} 
 & Client 1 & Client 2 & Client 3 & Client 4 & Client 5 & Client 6 & Client 7 & Client 8 & Client 9 \\ \hline
Centralized training & \multicolumn{3}{c|}{Centralized test result: 56.7} & \multicolumn{3}{c|}{Centralized test result: 58.0} & \multicolumn{3}{c}{Centralized test result: 52.3} \\ \hline
Local   training & 46.1 & 50.2 & 49.6 & 50.2 & 49.6 & 50.0 & 40.62 & 46.33 & 37.10 \\ \hline
FedAvg & 44.1 & 48.3 & 47.1 & 44.2 & 40.5 & 42.6 & 35.6 & 38.1 & 34.2 \\
FedProx & 44.5 & 48.3 & 46.6 & 48.0 & 46.3 & 43.9 & 34.2 & 37.5 & 34.6 \\
FedAdam & 45.6 & 49.6 & 50.2 & 50.1 & 51.2 & 49.0 & 42.9 & 49.5 & 42.8 \\
FedDPA & 47.3 & 50.2 & 50.6 & 52.3 & 51.9 & 50.4 & 43.5 & 49.6 & 43.2 \\
Shepherd & 43.4 & 45.0 & 46.2 & 37.6 & 36.8 & 39.5 & 35.6 & 35.3 & 30.7 \\ \hline
\textbf{Pilot} & \textbf{49.2} & \textbf{52.8} & \textbf{52.3} & \textbf{54.7} & \textbf{53.2} & \textbf{53.7} & \textbf{45.7} & \textbf{50.2} & \textbf{44.9}\\ \hline
\end{tabular}
\caption{Comparison with state-of-the-art methods under FL-oriented general VQA scenario.}
\label{result2}
\vspace{-1.0em}
\end{table*}

\subsection{Baselines}
We compare our framework Pilot with 5 state-of-the-art FL algorithms: FedAVG~\shortcite{mcmahan2017communication}, FedProx~\shortcite{li2020federated}, FedAdam~\shortcite{reddi2020adaptive}, Shepherd~\shortcite{zhang2024towards}, and FedDPA~\shortcite{yang2024dual}.
FedAvg takes the weighted average of all training parameters as a standard optimization method.
FedProx focuses on local model correction, and FedAdam focuses on introducing momentum on the server side to stabilize global model updates.
Shepherd and FedDPA are FedIT task method.
We also show local training and centralized training as references,
where local training is trained by using one client's dataset without collaboration. 
Centralized training is training all datasets centrally.

\subsection{Implementation Details}
In all experiments, we use LLaVA 1.5~\cite{liu2024visual} as multimodal Large Language Model.
LLaVA utilizes the pre-trained CLIP visual encoder ViT-L/14~\cite{radford2021learning} to extract visual features, and LLM utilizes Vicuna-V1.5-7b~\cite{chiang2023vicuna}.
The client-specific adapter and task-specific adapter parameters in stage 1 are initialized by the pre-trained MLP connector parameters in LLaVA.
All cross-task adapter parameters in stage 2 are initialized by the local task-specific adapter obtained in stage 1.
For all client text-adapter, we use the LoRA~\cite{hu2021lora} parameter efficient tuning technique to train LLM.
The rank of LoRA is 64 with a scalar $\alpha=128$.
During instruction tuning, we only fine-tune the parameters of the connector and LoRA while keeping the rest of the LLM frozen.
The learning rates of stage 1 and stage 2 are set to 2e-5 and 4e-5, respectively.
The number of text-adapter parameter aggregates Top-$M$ is set to 6
The coefficients of $\lambda_0$, $\lambda_1$, and $\lambda_2$ are 0.1, 0.1, and 0.01, respectively. 
The number of local cycles $E$ is set to 1 and the number of communication rounds $R$ is 3.
We train Pilot on 8 A100 GPUs (40G), with an effective batch size of 16 per GPU.
To ensure fairness, for all baselines, all additional hyperparameters involved in the compared methods use their best settings.

\subsection{Experimental Results}
For the VQA task (including ScienceQA, GQA, and OCRVQA), we calculate the accuracy of predicting answers against ground truth.
For the caption task, we report the CIDEr score.
For the grounding task, we employ Intersection-over-Union (loU) as the evaluation metric.
Specifically, a prediction is deemed accurate only when its loU exceeds or equals 0.5.
%
Table~\ref{result1} and Table~\ref{result2} show the comparison of Pilot with other five methods in the federated scenarios of visual understanding and General VQA. 
%
Our framework outperforms all baselines in two task scenarios.
We found that the performance of FedAVG method is lower than local training.
This shows that the heterogeneity of multimodal tasks leads to greater parameter conflicts, and simple aggregation has a negative impact on local models.
At the same time, we observed that FedAvg method is better than Shepherd.
The former aggregates all training parameters, while the latter only aggregates LLM training parameters.
This result also indirectly reflects the importance of visual information to the FedMIT task.
Compared with FedDPA, only improving the LLM training parameters does not achieve the desired effect,
which also illustrates the need to consider the differences between different tasks and the necessity of learning general knowledge from other tasks.
Finally, the results prove that our method can not only overcome the heterogeneity between tasks, but also collaborate with all clients to improve the performance of local models.

\subsection{Ablation Studies}
Here, we show the results of several variants of our method in the FL-oriented visual understanding scenario to demonstrate the effectiveness of the main modules of our method.
We first analyze the impact of the four components of our framework (i.e., cross-task adapter, difference loss, auxiliary loss, and adaptive text-adapter aggregation (ATA)) on the model performance, and further evaluate the effectiveness of the proposed method.
Table~\ref{ablation} shows the average score of all clients for the same task.
We found that without using the above methods, the performance of our method is lower than that of local training, which does not have the ability to overcome task heterogeneity.
Then we add adaptive text-adapter aggregation, and the performance of our method is improved and outperforms local training, which demonstrates that this module can effectively alleviate the impact of task heterogeneity.
When we add the cross-task adapter, we  observe that the model performance improves on all clients.
Through auxiliary loss optimization, the performance of our framework can be further improved. 
The results show that the CT-MOA module is able to learn general knowledge from other tasks to improve model performance and cross-task capabilities.
Removing the difference loss, our connector has only one and no longer distinguishes between client-specific adapter and task-specific adapter.
The performance of the model has decreased, indicating that it is necessary to maintain the personalized information of the client.
The above results demonstrate the importance of each component in our method.

\begin{table}[]
\centering
\renewcommand{\arraystretch}{1.1}
\small
\setlength{\tabcolsep}{5.7pt}
\captionsetup{font={normalsize}}
\begin{tabular}{cccc|c|c|c}
\hline
\multicolumn{1}{c|}{\multirow{2}{*}{\textbf{ATA}}} & \multicolumn{1}{c|}{\multirow{2}{*}{\textbf{${\psi}^c$}}} & \multicolumn{2}{c|}{\textbf{Loss}} & \multirow{2}{*}{\textbf{GQA}} & \multirow{2}{*}{\textbf{COCO}} & \multirow{2}{*}{\textbf{RefCOCO}} \\ \cline{3-4}
\multicolumn{1}{c|}{} & \multicolumn{1}{c|}{} & $\mathcal{L}_{b}+\mathcal{L}_{z}$ & $\mathcal{L}_{d}$ &  &  &  \\ \hline
$\times$ & $\times$ & $\times$ & $\times$ & 46.9 & 105.7 & 30.6 \\
$\checkmark$ & $\times$ & $\times$ & $\times$ & 47.5 & 110.6 & 39.8 \\
$\checkmark$ & $\checkmark$ & $\times$ & $\times$ & 49.3 & 114.5 & 47.5 \\
$\checkmark$ & $\checkmark$ & $\checkmark$ & $\times$ & 49.8 & 120.7 & 48.2 \\ \hline
$\checkmark$ & $\checkmark$ & $\checkmark$ & $\checkmark$ & \textbf{50.4} & \textbf{124.0} & \textbf{51.0}\\ \hline
\end{tabular}
\caption{Ablation studies under FL-oriented visual understanding scenario.}
\vspace{-1.1em}
\label{ablation}
\end{table}

\subsection{Further Remarks}
\textbf{Building Cross-task CLIP.}
Our method modifies the MLLM connector to learn general knowledge from other tasks to improve model performance and cross-task capabilities.
But the natural thought is: why not use CLIP?
To answer this question, we unfreeze CLIP and train all MLP layers in CLIP in the same way as the connector, called CT-CLIP.
We conduct experiments on the visual understanding scenario for federated learning.
Table~\ref{clip} shows the average scores of all clients for the same task,
where AP denotes local model Activation Params and CP denotes Communication Params.
Experimental results show that for the FedMIT task, learning cross-task visual information from different tasks is an effective solution.
We observe that although CT-CLIP outperforms the Pilot, it comes at the expense of additional training parameters and communication parameters.
Compared with Pilot, the communication parameters sent to the server increase by 0.2B, and the local client activation parameters increase by 1.25B.
With the increase of tasks, the computational cost is unacceptable.
Therefore, we give priority to the more simple and efficient method.

\begin{table}[]
\renewcommand{\arraystretch}{1.2}
\centering
\small
\setlength{\tabcolsep}{5.8pt}
\captionsetup{font={normalsize}}
\begin{tabular}{c|c|c|c|c|c}
\hline
\textbf{Methods} & \textbf{GQA} & \textbf{COCO} & \textbf{RefCOCO} & \textbf{AP} & \textbf{CP} \\ \hline
Pilot & 50.4 & 124.0 & 51.0 & \textbf{0.5B} & \textbf{0.3B} \\ \hline
+ CT-CLIP & \textbf{51.6} & \textbf{126.1} & \textbf{52.7} & 1.75B & 0.5B \\ \hline
\end{tabular}
\caption{Impact of vision encoder improvements on model performance.}
\label{clip}
\vspace{-0.8em}
\end{table}

\begin{table}[]
\centering
\renewcommand{\arraystretch}{1.1}
\small
\setlength{\tabcolsep}{10pt}
\begin{tabular}{c|c|c|c}
\hline
\textbf{Selection Strategy} & \textbf{GQA} & \textbf{COCO} & \textbf{RefCOCO} \\ \hline
Same   task client & 49.6 & 120.2 & 50.3 \\
All   clients & 47.9 & 109.7 & 45.2 \\
Pilot ($M$=5) & 49.7 & 122.8 & 50.6 \\
Pilot ($M$=6) & \textbf{50.4} & \textbf{124.0} & 51.0 \\
Pilot ($M$=7) & 50.3 & 123.2 & \textbf{51.7} \\ \hline
\end{tabular}
\caption{Compare different text-adapter parameter aggregation strategies.}
\label{text-adapter}
\vspace{-0.5em}
\end{table}

\begin{table}[!h]
\centering
\renewcommand{\arraystretch}{1.1}
\small
\setlength{\tabcolsep}{8pt}
\begin{tabular}{c|c|c|c}
\hline
\textbf{Initialization Strategy} & \textbf{GQA} & \textbf{COCO} & \textbf{RefCOCO} \\ \hline
Random & 50.1 & 122.7 & 50.3 \\
\textbf{Pilot} & \textbf{50.4} & \textbf{124.0} & \textbf{51.0} \\ \hline
\end{tabular}
\caption{Compare different cross-task adapter parameters initialization strategies.}
\label{initialization}
\vspace{-1.1em}
\end{table}

\textbf{Different Text-adapter Parameter Aggregation Strategies.}
In our framework, we adopt the adaptive text-adapter aggregation method.
For all text adapter parameters, we adopt a weighted optimization aggregation strategy based on euclidean distance and select the Top-$M$ parameters for weighted averaging.
%
%
We compared different aggregation strategies, such as aggregating only parameters of clients with the same task (same task client), aggregating parameters of all clients (all clients), and different Top-$M$ selections.
As shown in Table~\ref{text-adapter}, due to the heterogeneity between tasks, aggregating on all client will lead to parameters conflicts.
In addition, aggregating only on the same task is a suboptimal solution because it does not utilize the semantic knowledge of other tasks.
At the same time, we tested the impact of different top-$M$ on model performance, and the results showed that the adopted adaptive aggregation method can benefit from positive influences while reducing negative interference.

\textbf{Cross-task Adapter Initialization Strategy.}
In our framework, the role of the cross-task adapter is to extract cross-task knowledge.
We compared initialization with task-specific adapter parameters with training from scratch (random).
As shown in Table~\ref{initialization}, training from scratch leads to performance degradation for all tasks.
Using task-specific adapter parameters initialization provides a good starting point for the module and can help the local client better extract cross-task knowledge.

\section{Conclusion}
%
In this paper, we propose a federated multimodal instruction tuning framework to solve the new task of federated multimodal instruction tuning by collaboratively utilizing distributed data from different local clients to learn cross-task knowledge without being affected by task heterogeneity during instruction tuning.
Through two stages ``adapter on adapter" strategy, our model can capture the personalized information of local data and the task-related multimodal information, and can also adapt to the differences between different tasks.
Our method achieves state-of-the-art results in two cross-task scenarios.
%

\section{Acknowledgments}
This work was supported by National Natural Science Foundation of China under Grants 62036012, U23A20387, 62322212, 62072455, in part by Pengcheng Laboratory Research Projectunder Grant PCL2023A08, in part by Alibaba Innovative Research Program, and also in part by CAS Projectfor Young Scientists in Basic Research(YSBR-116).

\bibliography{aaai25}

\end{document}